\documentclass[preprint,journal]{vgtc}               

\ifpdf
  \pdfoutput=1\relax                   
  \usepackage{graphicx}                
  \DeclareGraphicsExtensions{.pdf,.png,.jpg,.jpeg} 
\else
  \usepackage{graphicx}                
  \DeclareGraphicsExtensions{.eps}     
\fi%

\graphicspath{{figs/}{pictures/}{images/}{./}} 

\usepackage{amsmath,amssymb}
\usepackage{microtype}                 
\PassOptionsToPackage{warn}{textcomp}  
\usepackage{textcomp}                  
\usepackage{mathptmx}                  
\usepackage{times}                     
\usepackage{cite}                      
\usepackage[compact]{titlesec}  
\titlespacing{\section}{0pt}{5pt}{2pt}
\titlespacing{\subsection}{0pt}{4pt}{2pt}

\newcommand{\Rspace}{\mathbb{R}}
\newcommand{\Mspace}{\mathbb{M}}
\newcommand{\Xspace}{\mathbb{X}}
\newcommand{\Sspace}{\mathbb{S}}
\newcommand{\Ispace}{\mathbb{I}}
\newcommand{\Lspace}{\mathbb{L}}
\newcommand{\Rerror}{\mathcal{R}}
\newcommand{\Ferror}{\mathcal{F}}
\newcommand{\Gerror}{\mathcal{G}}

\usepackage{enumitem}
\setitemize{noitemsep,topsep=1pt,parsep=1pt,partopsep=1pt}

\newcommand{\para}[1]{\vspace{1pt}\noindent\textbf{#1.}}

\onlineid{1243}

\vgtccategory{Application}

\vgtcinsertpkg
	
\title{Scalable Topological Data Analysis and Visualization for Evaluating Data-Driven Models in Scientific Applications}

 \author{
 Shusen Liu, Di Wang, Dan Maljovec, Rushil Anirudh, Jayaraman J. Thiagarajan, Sam Ade Jacobs \\
   Brian C. Van Essen, David Hysom, Jae-Seung Yeom, Jim Gaffney, Luc Peterson, Peter B. Robinson \\ Harsh Bhatia,  Valerio Pascucci,  Brian K. Spears and Peer-Timo Bremer} 
 \authorfooter{
 \vspace{-2mm}
 \item
Shusen Liu, Rushil Anirudh, Jayaraman J. Thiagarajan, Sam Ade Jacobs, Brian C. Van Essen, David Hysom, Jae-Seung Yeom, Jim Gaffney, Luc Peterson, Peter B. Robinson, Harsh Bhatia, Brian K. Spears, Peer-Timo Bremer are with Lawrence Livermore National Laboratory. E-mail: \{liu42, anirudh1, jayaraman1, jacobs32, vanessen1, hysom1, yeom2, gaffney3, peterson76, robinson96, bhatia4, spears9, bremer5\}@llnl.gov.
 \item
Di Wang, Dan Maljovec, Valerio Pascucci are with SCI Institute, University of Utah. E-mail: \{orpheus92, maljovec, pascucci\}@sci.utah.edu.}

\abstract{
With the rapid adoption of machine learning techniques for large-scale applications in science and engineering comes the convergence of two grand challenges in visualization. First, the utilization of black box models (e.g., deep neural networks) calls for advanced techniques in exploring and interpreting model behaviors. Second, the rapid growth in computing has produced enormous datasets that require techniques that can handle millions or more samples. Although some solutions to these interpretability challenges have been proposed, they typically do not scale beyond thousands of samples, nor do they provide the high-level intuition scientists are looking for. 
Here, we present the first scalable solution to explore and analyze high-dimensional functions often encountered in the scientific data analysis pipeline. By combining a new streaming neighborhood graph construction, the corresponding topology computation, and a novel data aggregation scheme, namely \textit{topology aware datacubes}, we enable interactive exploration of both the topological and the geometric aspect of high-dimensional data. Following two use cases from high-energy-density (HED) physics and computational biology, we demonstrate how these capabilities have led to crucial new insights in both applications. 
} 

\begin{document}


\firstsection{Introduction}

\maketitle

Driven by growing computing power and ever more sophisticated diagnostic instruments, many scientific applications have turned from being data poor to data rich. 
As a result, progress in these fields increasingly depends on our ability to understand and learn from data that are overwhelming in both size and richness. 
Scientists and analysts have turned to data-driven models, particularly those based on deep neural networks, to handle this increase in volume and complexity.
These models exploit the complex nonlinear relationships in such large-scale data to \textit{learn} effective predictive models.
They allow us to deal with multivariate and multimodal data (i.e., images, time-series, scalars, etc.) in ways inconceivable a decade ago.
However, these flexible models create new challenges, especially in scientific applications.
In commercial AI systems, sheer performance, measured by predictive power or classification accuracy, is paramount. 
Science and engineering practitioners require satisfactory performance, but they also demand insights and understanding. 
Yet, the very complexity that increases model performance often obscures our understanding and hence fails to provide new insights. 
%
%
This scenario has produced a growing need for techniques to interpret the results of data-driven models and to explore their training process and final error landscape to assess their reliability.


Despite a recent surge in techniques for model interpretation~\cite{SimonyanVedaldiZisserman2013, ZeilerFergus2014, YosinskiCluneNguyen2015, OlahMordvintsevSchubert2017}, most approaches focus on explaining a given prediction and often only for a specific problem domain, such as natural language processing or computer vision. 
Although these techniques are useful to understand individual predictions, they provide few insights into the overall behavior of a model or its internal representations. 

In this paper, we are concerned with two large-scale applications: a high-energy-density physics ensemble exploring inertial confinement fusion and a multiscale simulation of a human cell membrane. 
In the former application, some members of our team have built a sophisticated surrogate model to explore a large ensemble of simulations with up to 10 million members.
The latter is driven by a sampling process in a high-dimensional latent space, and scientists are interested in understanding and comparing sample distributions. 
In both cases the goal is not only to understand the gross simulation results, e.g., the dependence of fusion yield on design parameters, but also, more importantly, to provide insight and establish confidence in the model itself.
Therefore, we need techniques to evaluate the errors and uncertainties in the model with a special emphasis on their variation within the model domain. 
For example, complex physics models are known to have strong non-linearities that often lead to a rapidly changing response in a very small region of the input domain of the simulation (also referred to as the parameter space). 
This sharp nonlinear behavior can lead to models that exhibit low global errors and good model convergence, while at the same time producing wildly incorrect results in localized regions of parameter space that are often most interesting to the scientists. 
Moreover, scientists typically use these models and their predictions to design new experiments.
They need tools that can discriminate between regions of parameter space where the model has errors small enough to make the model useful for design work.  
Likewise, they need to know where errors have grown large enough to make the model unreliable for predicting real-world conditions.
Such requirement highlights the need in the scientific community to augment global loss functions and response metrics with methods that can steer the scientific user toward regions of high value and confidence.
Especially for the complex models considered in this paper, with multiple, interdependent loss functions and millions of training and evaluation samples, it is critical to understand the impact of different loss functions and how the errors might be correlated to the underlying physics.

%
 
None of the existing techniques are particularly suited for the analysis of these scientific models because they either do not scale to the required sample sizes or do not provide the necessary details. 
For example, due to the curse of dimensionality, it is not uncommon for ensembles, even with a large number of samples, to have a small number of salient observations~\cite{Liu18usc} that are crucial for interpretation, yet are not easily detected automatically. 
Preserving such observations creates conflicting objectives, as the data size demands aggregating information into global trends, yet doing so obscures small-scale details. 

As discussed in more detail in Section~\ref{sec:task}, many of the questions raised above can be expressed as the analysis of a high-dimensional function.
For scientific analysis, this function might be a quantity of interest, such as the yield, whereas for model interpretation and validation, it might be a loss function or another indicator of prediction quality. 
In general, we are given a high-dimensional \textit{domain}, i.e., a set of input parameters or a latent space, as well as a scalar function on that domain, to analyze. 
We propose to use topological techniques to address this challenge.
In general, computing the topology of a given function provides insights into both its global behavior and it local features, since the measure of importance is variation in its function value, i.e., persistence, rather than the size of a feature.
Furthermore, topology provides a convenient abstraction for both analysis and visualization, whose complexity depends entirely on the function itself rather than the number of samples used to represent it. 
However, topological information alone can provide only limited insight because, for example, knowing that there exist two significant modes is of little use without knowing where in the domain these modes exist and how much volume they account for. 
We, therefore, propose to augment the topological information with complementary geometric information by introducing the notion of topological datacubes.
These datacubes are sampled representations linked directly to the topological structure, which provide capabilities for join exploration among scatterplots, parallel coordinates, and topological features.
Finally, we introduce new streaming algorithms to compute both the topological information and the corresponding datacubes for datasets ranging from hundreds of thousands to eight million data points.
More specifically, we have

\begin{itemize}
    \item identified the shared analysis challenges posed by many data-driven modeling applications;
    \item developed a robust topological data analysis computation pipeline that scales to millions of samples;
    \item developed an interactive visual analytics system that leverages both topological and geometric information for model analysis;
    \item addressed the model analysis challenges of a data-driven inertial confinement fusion surrogate and solved the adaptive sampling evaluation problem for a large-scale biomedical simulation.

\end{itemize}

\section{Related Works}
Since we focus on addressing specific application challenges, there are limited works aiming to solve the exact problem.
Here we review several topics that relate to the individual components of the system or provide relevant background for the overall application. 

\para{Scientific Machine Learning}
Data analysis has always been one of the driving forces for scientific discovery. Many statistical tools, i.e., hypothesis testing~\cite{anderson1958introduction}, have co-developed with many scientific disciplines.
The application of modern machine learning tools in scientific research has long attracted scientists' attention~\cite{mjolsness2001machine}.
Machine learning has been successfully utilized for solving problems in a variety of fields, such as bioinformatics~\cite{baldi2001bioinformatics}, material science~\cite{butler2018machine}, and physics~\cite{peterson2017zonal}. In a recent high-energy-density physics application~\cite{peterson2017zonal}, a random forest regressor has been adopted for identifying a previously unknown optional configuration for the inertial confinement fusion experiments (more background on inertial confinement fusion is included in Section~\ref{sec:app_icf}).
With recent advances of powerful learners (i.e., deep neural network), coupled with the data analysis challenge, scientists are increasingly interested in leveraging these models for scientific discovery.

\para{Deep Learning Model Interpretation}
The opaque nature of deep neural networks has prompted many efforts to interpret them. 
In the machine learning community, many approaches have been proposed, notably for common architecture such as the convolution neural network (CNN) ~\cite{SimonyanVedaldiZisserman2013, ZeilerFergus2014, YosinskiCluneNguyen2015}, to probe into the mechanism of the model.
Similarly, visual analytics systems that focus on interactive explorations of model internals have also been introduced in the visualization community, for vision~\cite{liu2018analyzing}, text~\cite{liu2019nlize}, reinforcement learning~\cite{wang2019dqnviz}, and more~\cite{ming2019rulematrix}.
However, these model-specific techniques are closely tied to particular architectures or setups, making them less flexible to a variety of application scenarios. 
Alternatively, we can approach the interpretation challenge from a model agnostics perspective. Recently, a few studies~\cite{RibeiroSinghGuestrin2016, KrausePererNg2016, LundbergLee2017} have focused on building a general interpretation engine. For example, the LIME~\cite{RibeiroSinghGuestrin2016} explains a prediction by fitting a localized linear model that approximates the classification boundary around the given example.
Considering the high-dimensional nature of the internal state of the neural network models, we believe there is a unique opportunity to build a general purpose neural-network-interpreting tool by exploiting high-dimensional visualization techniques.
In this work, we demonstrate a firm step toward this goal, where we provide valuable diagnostic and validation capabilities for designing a surrogate model of a high-energy-density physics application.

\para{Topological Data Analysis}
Compared to traditional statistical analysis tools, topological data analysis employs quite different criteria that allow it to pick up potentially important outliers that would otherwise get ignored in standard statistical analysis, making topological data analysis tools uniquely suitable for many scientific applications.
The topology of scalar valued function has been utilized for analyzing scientific data in previous works~\cite{gyulassy2005topology, gyulassy2008practical, bennett2012combining}.
The scalability challenge for handling functions defined on the large 3D volumetric domain has also been addressed in several previous works, e.g., parallel merge tree computation~\cite{landge2014situ}.
However, in many scientific applications,  scientists are also interested in studying the properties defined in a high-dimensional domain.
To address such a challenge, the HDVis work~\cite{gerber2010visual} was introduced for computing the topology of a sampled high-dimensional scalar function. 
Unfortunately, the applications of the high-dimensional scalar topology we see so far contain only a relatively small number of samples. 
An apparent mismatch exists between the scalability of the existing high-dimensional scalar function topology implementation and the large datasets for our target applications. 
This work aims to fill the gap by addressing both the computational and visualization challenges.
For background on topological data analysis, please refer to Section~\ref{sec:topology}.

\para{Data Aggregation Visualization}
As the number of samples increases in a dataset, a visual analytics system not only needs to cope with the computational/rendering strain but also to handle the visual encoding challenges. For example, if we simply plot millions of points in a parallel coordinate or scatterplot, besides the speed consideration (i.e., drawing each point as an SVG object using \emph{d3.js} will not be ideal), the occlusion and crowding could potentially eliminate any possibility of extracting meaningful information from the data. Many previous works in visualization have been proposed to address these scalability challenges, ranging from designing novel visual encoding (e.g., splatterplot~\cite{mayorga2013splatterplots}, stacking element plot~\cite{dang2010stacking}) to directly modeling and rendering the data distribution (e.g., Nanocubes~\cite{lins2013nanocubes}, imMens~\cite{liu2013immens}, density-based parallel coordinate~\cite{artero2004uncovering}).
However, most existing approaches aggregate information globally or according to certain geological indexing for faster queries. 
In this work, the proposed system combines visualization components for both topological and geometric features. Therefore, in order to scale the linked view visualization system beyond millions of points, we adopt a topology-aware datacube design for aggregating large data according to their topological partition.


\section{Application Tasks Analysis}
\label{sec:task}
This section provides the application background and identifies: (i) the analysis tasks shared by many data-driven applications; (ii) why these tasks are important; and (iii) how we can solve these tasks by combining topological data analysis with interactive data visualization.
The system has been developed jointly through a continuing collaboration with two application teams in high-energy physics and computational biology, respectively. 
In both cases we are tasked with solving the analysis challenges encountered throughout the processing pipeline, including sample acquisition, sampling, modeling, and analysis.
Despite the disparate application domains, we have encountered a number of recurring analysis tasks, which often lie at the very heart of the scientific interpretation.
This has motivated the development of dedicated visual analysis tool to streamline the analysis process and accelerate discovery.

Data is at the center of any analysis task. 
Experimental data is typically considered the gold standard, but it is often too costly or time consuming to perform all desired experiments, and/or the phenomenon in question cannot be directly observed. 
In these cases, computer simulations at various fidelities have become indispensable to plan and interpret observations. 
However, such simulations are rarely completely accurate and often rely on educated guesses of parameters or known approximations of physical processes.
To deal with such uncertainties and calibrate the simulations, scientists turn to simulation ensembles in which the same (computational) experiment is simulated thousands or millions of times with varying parameters, initial conditions, etc.
The corresponding computational pipeline typically has three main stages: sampling, simulation, and modeling.
We start by generating samples in the input parameter space.
We then run simulations on all input parameter combinations and gather the outputs to create the ensemble.  
Finally, in the modeling stage, the simulation results are used to train a cheaper surrogate for one or multiple of the output quantities to enable statistical inference, uncertainty quantification, or parameter calibration. 
Both the sampling and the modeling stage can benefit significantly from visual analytics.
In particular, we have identified four generic tasks we typically encounter independent of the specific application:


\begin{itemize}
  \item \textbf{T1:} Analyze the sampling pattern to ensure uniform coverage or verify a sampling objective
  \item \textbf{T2:} Explore quantities of interest with respect to the input parameter space
  \item \textbf{T3:} Verify model convergence, explore residual errors, and evaluate local and global reliability
  \item \textbf{T4:} Explore the sensitivities of the model with respect to the input parameters
\end{itemize}

How we sample the high-dimensional input space has a significant impact on the downstream analysis.
Depending on the nature of the application, we may have different preferences in the sample pattern and thus need to verify whether the sampling pattern satisfies the required properties (\textbf{T1}).
Given the simulation outputs, we then need to identify where we achieved or failed to achieve our objective, i.e., induce nuclear fusion, how stable successful solutions are, etc.
Typically, such process requires us to explore some quantity of interest, i.e., energy yield of the physical simulation, in the high-dimensional input parameter space (\textbf{T2}). 
Once we obtain the simulation and build a surrogate model, we need to be able to evaluate the model behavior and interpret the model's internal representations (\textbf{T3}).
The \textbf{T2}, \textbf{T3} focus on identifying regions in the high-dimensional space corresponding to certain meaningful measures.
As discussed in the introduction, topological data analysis that identifies local peaks of a function can be an effective tool for discovering and exploring these regions in a multiscale manner. 
However, due to the high-dimensional nature of the space, a large number of samples are often required to provide adequate coverage of the space.
As a result, we need to make sure the topological data analysis computation pipeline can reliably scale to large sample sizes (i.e., beyond tens of millions of points). We address the scalability challenges by adopting a streaming computation pipeline (discussed in in Section~\ref{sec:topology}).

Aside from studying the behavior of functions in certain high-dimensional space (\textbf{T2, T3}), we also want to understand the relationship between specific input parameters and model output (\textbf{T4}), for example, judge sensitivities.
A global regression will often not yield the desired result, because the relationships can be both complex and highly localized.
An alternative is to leverage the topology-guided partition (\textbf{T2, T3}) and examine the localized and potentially simpler trend. 
However, the topological data analysis does not really capture any geometric structure nor can it help reason about individual dimensions.
Therefore, to fully utilize the revealed topological information, we need to provide complementary geometric information.  In the proposed system, we adopted scatterplots and parallel coordinates, which intuitively encode data dimensions and support flexible selection operations that benefit from the linked view visual exploration.
In order to scale the combined topological and geometrical explorations for large sample size and support interactive query with respect to topological feature (i.e., different extrema), 
we devised a topology-aware datacube to enable the interactively linked exploration between topological features and the corresponding samples' geometric information (see details in Section~\ref{sec:vis}).
 



\section{Streaming Topological Data Analysis}
\label{sec:topology}

As discussed above, the exploration and analysis introduced in this work rely on
high-dimensional topology and in particular high-dimensional extremum
graphs~\cite{CorreaLindstromBremer2011,Thomas13tvcg}.
Topological information provides insight into the overall structure of high-dimensional
functions and is a convenient handle for per-extremum localized analysis.
Here, we first introduce the necessary background in Morse theory and review
the original algorithm to approximate extremum graphs for sampled high-dimensional functions.
We then discuss a new streaming algorithm to compute extremum graphs and the corresponding streaming neighborhood graph construction approach, 
which mainly aim to overcome the memory bottleneck that hinders the scalability of the existing implementation.

\subsection{Morse Complex and Extremum Graphs}
\label{sec:standardAlgorithm}

%
Let $\Mspace$ be a smooth, $d$-dimensional manifold and
$f: \Mspace \rightarrow \Rspace$ be a smooth scalar function with gradient
$\nabla f$.
A \textit{critical point} is any point $x \in \Mspace$ where $\nabla f(x)=0$,
and all other points are \textit{regular}.
Starting from a regular point $p \in \Mspace$ and following the gradient forward and backward traces an \textit{integral} line that begins at a local minimum and ends at a local maximum.
The set of points whose integral line ends at a critical point defines the stable/unstable manifold for that point, and the \textit{Morse complex} is defined as the collection of all stable manifolds.
For dimensions beyond three, computing the entire Morse complex is infeasible~\cite{Gerber10tvcg}, so we follow~\cite{Bremer06Banff,CorreaLindstromBremer2011} and concentrate on extremum graphs.
An extremum graph contains all local maxima of the function together with the saddles connecting them.
To approximate the extremum graph for sampled functions, we define an undirected neighborhood graph on all input vertices to approximate $\Mspace$.
Given a graph, we approximate the gradient as the steepest ascending edge incident to a given vertex and construct the ascending integral line by successively following steepest edges to a local maximum.
In practice, this traversal is implemented as a short-cut union-find at near linear complexity per vertex.
In this process, each vertex is labeled by the index of its corresponding stable manifold, and saddles are identified as the highest vertex connecting two neighboring maxima~\cite{Gerber10tvcg}.
Subsequently, we form the extremum graph by considering arcs between saddles and neighboring maxima. 

\begin{figure}[htbp]
  \centering
   \vspace{-2mm}
     \includegraphics[width=.49\linewidth]{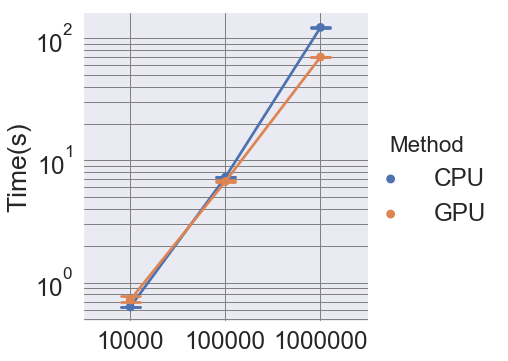}
    \includegraphics[width=.48 \linewidth]{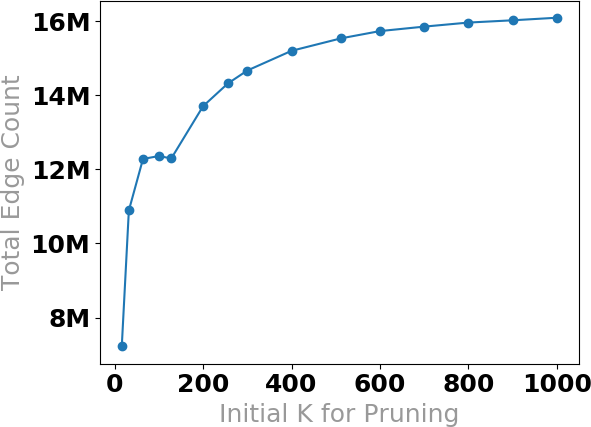}
    \vspace{-3mm}
   \caption{
   Left: the performance comparison between the CPU baseline streaming implementation and the CUDA accelerated version (5D sample with varying sizes).
   Right: a plot showing the total edges extracted from a symmetric,
   relaxed Gabriel graph computed on the same dataset as a function of the $k$
   used in the initial $k$-nearest neighbor graph. 
   }
  \label{fig:graph_degree}
  \vspace{-4mm}
\end{figure}



\subsection{Streaming Extremum Graph}
\label{sec:streamingAlgorithm}

One major road block for scaling the existing algorithm to large point count is the size of the neighborhood graph.
As the dimension of $\Mspace$ increases, more vertices are required to reliably approximate $f$. 
To identify the correct edges for topological computation, we often need to build neighborhood graphs that require several magnitude more storage space than the vertex alone, which may quickly reach the memory limitation of most desktop systems.

To address this challenge, we present a two-pass streaming algorithm for constructing extremum graphs that store only the vertices and an appropriate neighborhood look-up structure to avoid keeping the massive neighborhood graph in memory.
We can then obtain the necessary neighboring information on the fly to construct edges for each vertex. A vertex always maintains its currently steepest neighbor, which is initialized to be the vertex itself.
Once all edges have been seen, a near-linear union-find is used as a short-cut to the steepest neighbor link to point to the corresponding local maximum of each vertex, thereby constructing the sampled Morse complex.
Even though this structure can be constructed in a single pass, note that the identification of saddles and the subsequent cancellation of saddles and maxima also require the neighborhood information.
Consequently, we reiterate the steaming graph in a second pass to assemble the complex.
%

To support streaming extremum graph computation, we introduce a new approach to compute neighborhood graphs in a streaming manner.
For densely sampled space (i.e., sampling of simulation input parameters), Correa et al.~\cite{CorreaLindstrom2011} have demonstrated that $\beta$-skeletons and their relaxed variants provide significantly more stable results for computing topological structure~\cite{CorreaLindstrom2011}, in which an approximated $k$-nearest neighbor graph of sufficiently large $k$ is computed first and then pruned using the empty region test defined by $\beta$.
In this work, we build upon their work~\cite{CorreaLindstrom2011} and employ a streaming scheme to avoid storing the full graph in memory by doing neighborhood lookup and edge pruning for each point individually. We then store the steepest direction for topology computation. 
In our implementation, we utilize GPU to exploit the parallelism in the neighborhood query and the edge-pruning steps.
A comparison between the baseline CPU implementation and the GPU accelerated version is shown in the left plot of Fig.~\ref{fig:graph_degree} on a test function with a 5D domain and varying sample sizes (note the log scale in the y-axis, for the 1M case, the GPU version is approximately $70\%$ faster than the CPU counterpart).

To determine a suitable $k$, as illustrated in the right plot of Fig.~\ref{fig:graph_degree}, 
we gradually increase $k$ in the initial $k$-nearest neighborhood query stage and observe at which point the number of true edges of the empty region graph begins to stabilize.
We can see, in 5D space, that the curve starts to level off beyond $k\sim500$, indicating that adding more $k$ neighbors will not result in many more edges being discovered in the pruned graph. 
Note that the saturation point will vary based on the distribution of the data in the domain; the results are shown for data drawn from a uniform random distribution.

\section{Data Aggregation and Visualization}
\label{sec:vis}
%
We have described how to extend the scalability of the topological data analysis to handle large datasets.
Despite its strength, topological data analysis alone reveals little geometric information outside the location of critical points in parameter space. 
However, this information can be crucial to interpret data, and thus we propose a joint analysis of topological features and their corresponding geometric data as expressed through parallel coordinates and scatterplots. 
Here, we introduce \emph{topology-aware datacubes}, which not only aggregate data to enable interactive visual exploration but also maintain the topological feature hierarchy that allows interactive linked view exploration. 
As illustrated in Fig.~\ref{fig:interface}, the overall visualization interface (built on top of an existing high-dimensional data exploration framework $ND^{2}AV$~\cite{nddav}) consists of three views of the dataset: topological spines (a), scatterplots (b), and parallel coordinates plots (c). 
The topological spine shows the relative distance and size of the peaks in the function of interests. During an exploration session, the user can first assess the global relationship presented in the topological spine and parallel coordinate, and then focus on individual peaks of the function in the topological spine by selecting the local extrema, which will trigger updates in the parallel coordinates plot and scatterplot that reveal the samples corresponding to that extrema. 
By examining the parallel coordinates plot / scatterplot patterns that correspond to different selected extrema, we are able to discern the locations of the peaks in the function that make them different.

%

\subsection{Topology-Aware Data Cubes}
When dealing with large datasets, visualization systems often need to address the scalability challenge from two aspects. On one hand, as the number of sample increases, the standard visual encodings, such as scatterplots and parallel coordinates, become increasingly ineffective due to overplotting and thus overwhelm users' perceptual capacity. 
On the other hand, the increasing data size induces a heavy computational cost for processing and rendering, which may create latency that hinders the interactivity of the tool.
One strategy to address this problem is datacubes (OLAP-cubes) style aggregation techniques~\cite{lins2013nanocubes, liu2013immens}, 
which provide summary information (i.e., density/histogram) to overcome the overplotting while preserving joint information to enable linked selection and interactive exploration. 
However, such aggregation techniques are not directly applicable here as they summarize the dataset's entire domain without considering the topological segmentation.
Instead, we aim to aggregate information with respect to the topological structure of the data because the ability to interactively explore the topological hierarchy at different granularities is central to our analysis task.
Therefore, to achieve the design goal, the data aggregation structure should allow efficient query on different topological partitions of the data, which motivates us to introduce a topology-aware datacube design.

\begin{figure}[htbp]
\vspace{-2mm}
\centering
  \includegraphics[width=0.99\linewidth]{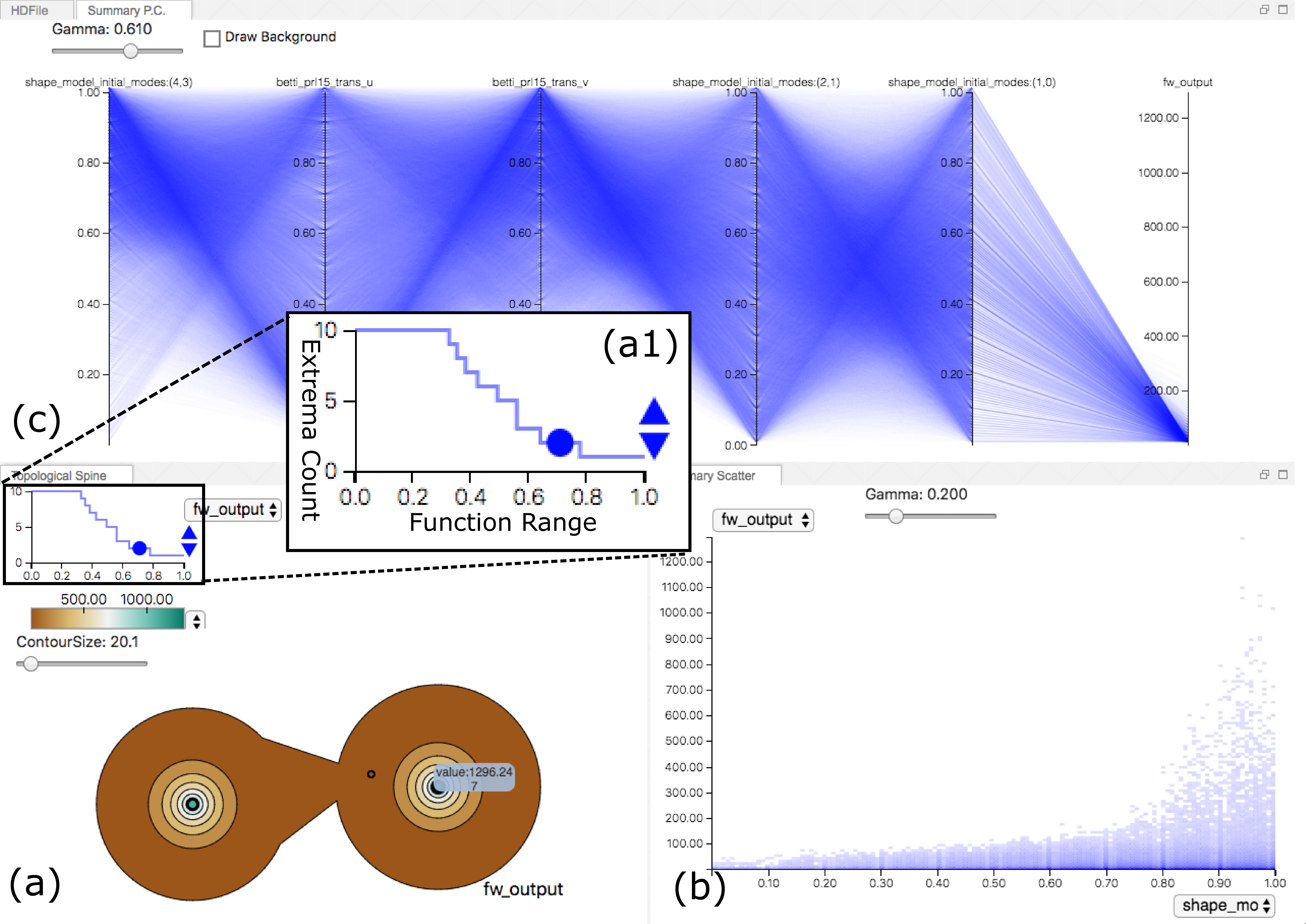}
  \vspace{-4mm}
 \caption{The proposed visualization interface consists of three views: topological spine (a), density scatterplot (b), density parallel coordinates plot (c). 
These views provide complementary information, and the linked selection enables a joint analysis of both geometric and topological features. In (a1), we show the persistence plot, which controls the number of extrema displayed in the topological spine.}
\label{fig:interface}
\vspace{-0.1in}
\end{figure}

%
%
The extremum graph (Section~\ref{sec:topology}) can be simplified by merging extrema (each corresponding to a segment of the data) and removing less significant saddles based on the \emph{persistence} value (i.e., the function value difference between the saddle and extremum pair: the higher the value, the more significant the topological feature). 
Since we often do not care about extrema with low persistence (i.e., corresponding to a less topologically significant structure), 
we can presimplify the topological segmentation hierarchy and focus only on the top levels (i.e., up to a dozen segments) for our analysis.
During the exploration, the user can explore different levels of granularity by altering the persistence threshold. 
In Fig.~\ref{fig:interface}(a1), the number of peaks is shown (y-axis) for a given persistence value (x-axis). 
The persistence values at the plateau regions (longer horizontal line segments in the diagram) signify more stable topological structures. A suitable persistence value is the one that can produce significant and stable topological structures.

For the topology-aware datacubes, we precomputed datacubes for samples in each leaf segment in the simplified topology hierarchy, where each datacube preserves the localized geometric features.
The segments with lower persistence will be merged into a topologically more significant partition of the data.   
Since we already precomputed the datacubes for each leaf segment, we can generate summary information for any higher level partition by aggregating the summary data in leaf segments on the fly.
%
%
%
%
Despite the effectiveness of the datacube for range queries, storing a full joint distribution in the datacube form can be extremely expensive. 
For our intended exploration and interaction (displaying parallel coordinates plots and scatterplots), we do not need access to all the joint distribution information.
Therefore, we compute and store lower dimensional data (up to three dimensions) cubes to reduce the storage while still supporting the interactive visualization. 
%

\begin{figure}[htbp]
\vspace{-1mm}
\centering
  \includegraphics[width=0.95\linewidth]{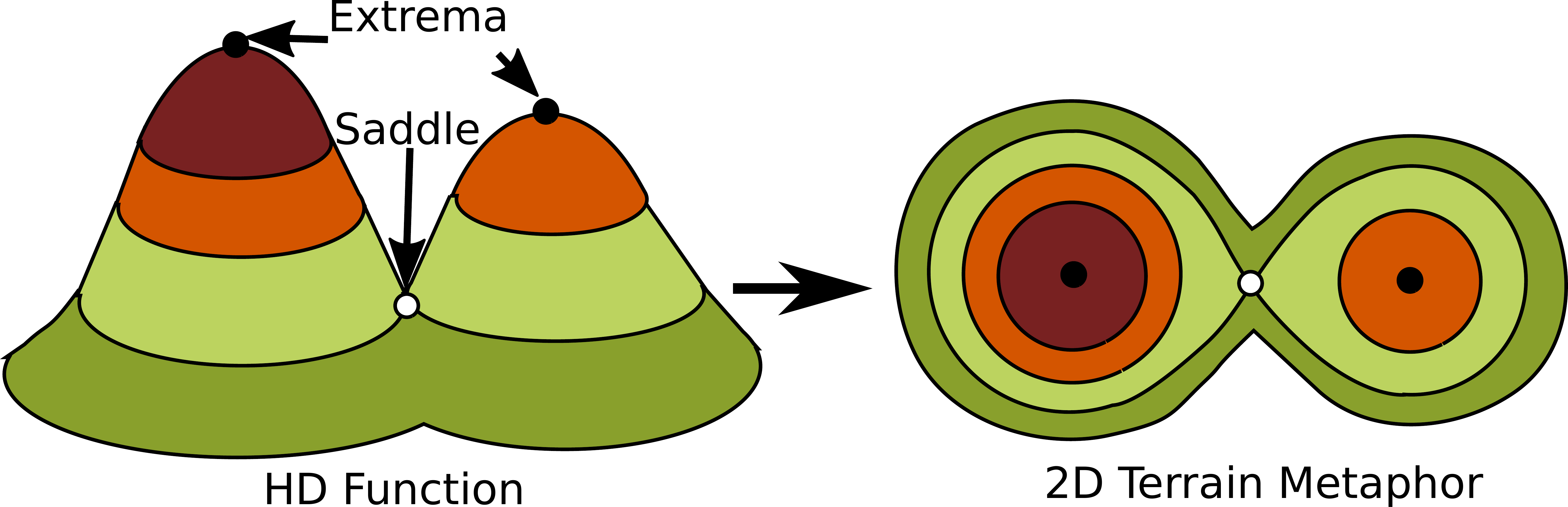}
  \vspace{-3mm}
 \caption{Topological spine. A 2D terrain metaphor of high-dimensional topological information~\cite{CorreaLindstromBremer2011}.}
\label{fig:topoSpine}
\vspace{-1mm}
\end{figure}

\para{Topological Spine}
The topological spine~\cite{CorreaLindstromBremer2011} view (see Fig.~\ref{fig:interface}(a)) visually encodes the extremum graph of a scalar field by showing the connectivity of the extrema and saddles together with the size of local peaks.
As illustrated in Fig.~\ref{fig:topoSpine}, the spine utilizes a terrain metaphor to illustrate the peaks (local extrema) and their relationships in the given function. 
Here, the contours around the extrema or across saddles indicate the level sets of the function, similar to the contour line in a topographic map. 
In~\cite{CorreaLindstromBremer2011}, the size of the contour $C_{f_i}$ is the number of samples above the contour function value $f_i$.
In our implementation, we computed the size as $\sqrt[d/2]{C_{f_i}}$ to better reflect the relative ``volume'' these points cover in the high-dimensional space ($d$ is the dimension of the function domain).
The layout of the critical points (the extrema and saddles) is a 2D approximation of the location of critical points in the high-dimensional domain. 
Since the function may have many small local structures, we simplify the extremum graph to focus on important topological features.
We consider a given local extremum less significant when the function value difference between the extremum and the nearby saddle (i.e., the persistence of the local extremum) is small. The extremum graph can be simplified by merging extrema with small persistence values.
As illustrated in Fig.~\ref{fig:interface}(a1), the number of extrema in the topological structure is controlled by the persistence plot, where the x-axis is the normalized function range, and the y-axis shows the number of local extrema at the current simplification level. 
We can identify the stable topological configuration by finding the widest plateau in the blue line along the x-axis.
Since the complexity of the spine does not depend on the number of samples used to define the function, the topological spine is a perfectly scalable visual metaphor that can easily be obtained from the precomputed datacubes instead of the raw samples.

%
%

%

\para{Density Scatterplots}
%
By querying the datacube, we can directly obtain the 2D histogram and render the estimated joint distribution density to avoid the overplotting issue in the standard scatterplot.
Due to the topology-aware datacube, the user can explore the density scatterplot (see Fig.~\ref{fig:interface}(b)) for any topological segments by aggregating leaf datacubes on the fly. 
%
%
To better visualize the potentially large dynamic range of the density value, we applied a gamma correction on the density value: $\alpha = Ad^{\gamma} $, where $\alpha$ denotes the opacity and $d$ denotes the density of each bin. %

\para{Density Parallel Coordinates}
%
Similarly, the parallel coordinates plot (see Fig.~\ref{fig:interface}(c)) can be drawn with selected 2D joint distributions from the datacube. 
To draw the density parallel coordinates plots, we first discretize each axis according to the resolution ($r$) of the datacube; thus, there would be $r$ bins on each axis. 
We then draw lines from each bin to every other bin on the adjacent axis; thus, there would be $r^2$ lines between every two adjacent axes. 
To draw each line, we query the corresponding density from the 2D joint distribution of the neighboring axes and map the value to the opacity (with gamma correction). 
Since every line between two adjacent axes requires only the information of corresponding dimensions, we need only the bivariate distribution of those two dimensions (or 3D if we want to support function range selection in PCP). Due to the discretization, the time to draw the parallel coordinates plot depends only on the resolution and the total number of dimensions.


%
%
%
%
%


\begin{figure}[htbp]
\centering
  \includegraphics[width=0.9\linewidth]{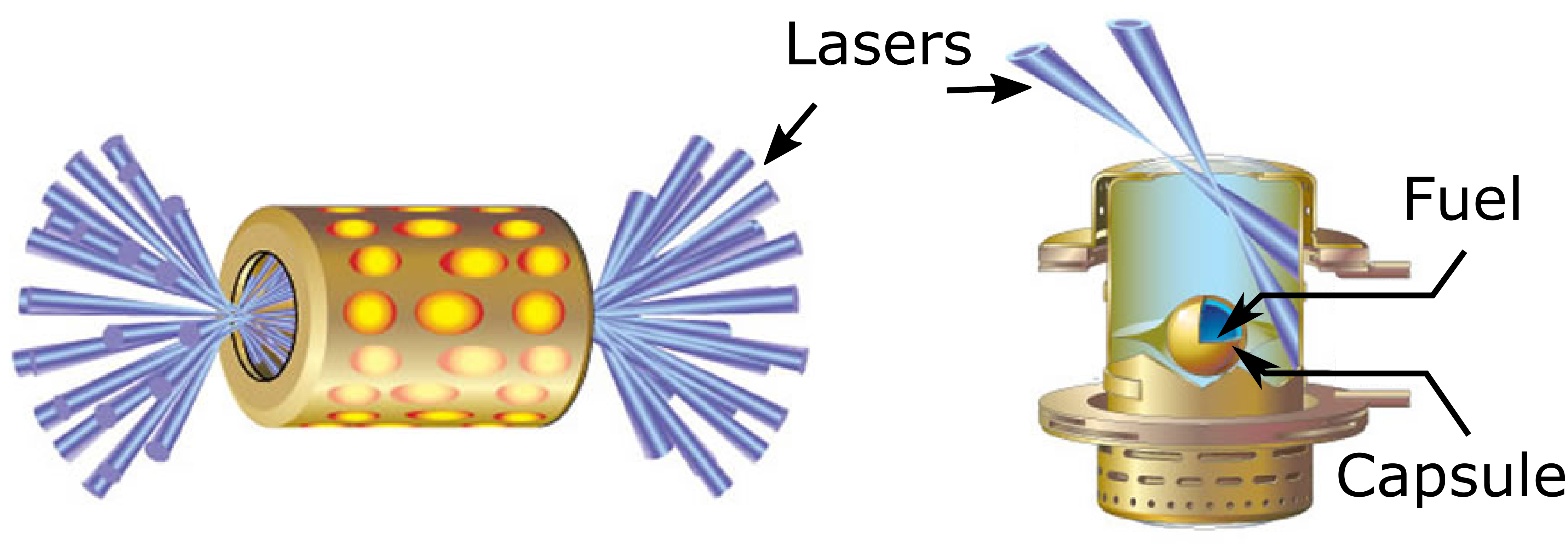}
  \vspace{-3mm}
 \caption{Inertial confinement fusion (ICF). Lasers heat and compress the target capsule containing fuel to initiate controlled fusion.
 }
\label{fig:ICF_capsule}
\vspace{-3mm}
\end{figure}

\section{Application: Inertial Confinement Fusion}
\label{sec:app_icf}

In this section, we will discuss the application of the proposed analysis framework for analyzing
simulations of inertial confinement fusion (ICF).
Controlled fusion has often been considered to be a highly effective energy source. At the National Ignition Facility (NIF) (at Lawrence Livermore National Laboratory),
one of the key objectives is to provide an experimental environment for controlled fusion research and thus laying the groundwork for using fusion as a clean and safe energy source.
Scientists utilize high-energy lasers to heat and compress a millimeter-scale target filled with frozen thermonuclear fusion fuel (see
Fig.~\ref{fig:ICF_capsule}). Under ideal conditions, the fusion of the fuel will produce enough neutrons and alpha particles (helium nuclei) for
the target to self-heat, eventually leading to a runaway implosion
process referred to as ignition. 
However, achieving an optimal condition by adjusting the target and laser parameters
in response to the experimental output (e.g., energy yield) is extremely challenging due to a variety of factors, such as the prohibitive cost of the actual physical experiment and incomplete diagnostic information.
Consequently, ICF scientists often turn to numerical simulations to help them
postulate the physical conditions that produced a given set of
experimental observations.

\begin{figure}[htbp]
\vspace{-2mm}
\centering
  \includegraphics[width=\linewidth]{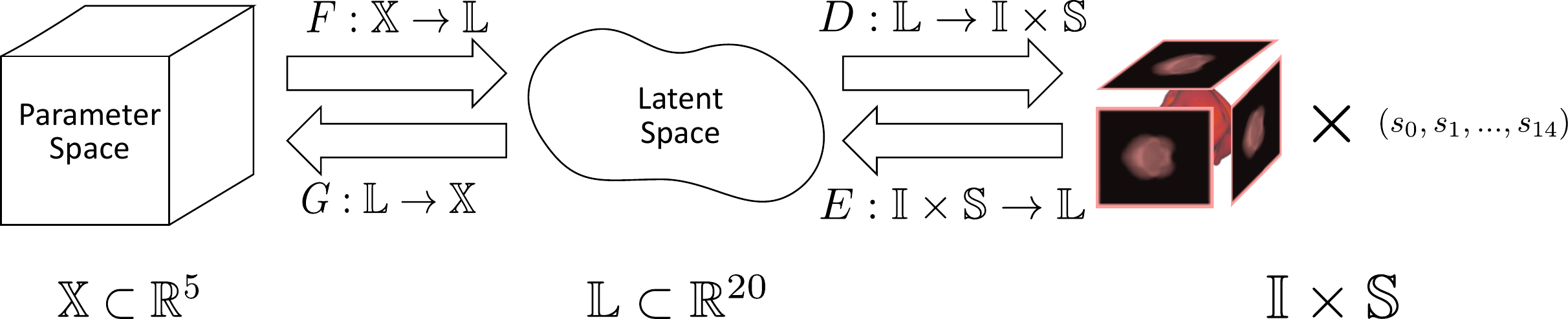}
	\vspace{-4mm}
 \caption{System diagram of the deep learning-based surrogate modeling used in the ICF study. A 5D input parameter space $\Xspace$ is mapped onto a
   20D latent space $\Lspace$, which is fed to a
   multimodal decoder $D$ returning the multiview X-ray images of the
   implosion as well as diagnostic scalar quantities.}
\label{fig:cycleGAN}
\vspace{-2mm}
\end{figure}

In this application, we focus on a large simulation ensemble (10M samples) produced by
a recently proposed semianalytic simulation model~\cite{gaffney2014thermodynamic, springer2013integrated}.
We capitalize on the abundance of simulations by building a complex multivariate
and multimodal surrogate based on deep learning architectures.
More specifically, the surrogate replicates outputs from the numerical physics model, including
an array of simulated X-ray images obtained from multiple lines of sight, as well as diagnostic scalar output quantities (see Fig.~\ref{fig:cycleGAN}).
As discussed in more detail below, this complex deep learning model is the first of its kind, at
least in this area of science. The model and the corresponding training dataset have been made publicly available~\cite{jagData}.

\begin{figure*}[!t]
\centering
  \includegraphics[width=0.97\linewidth]{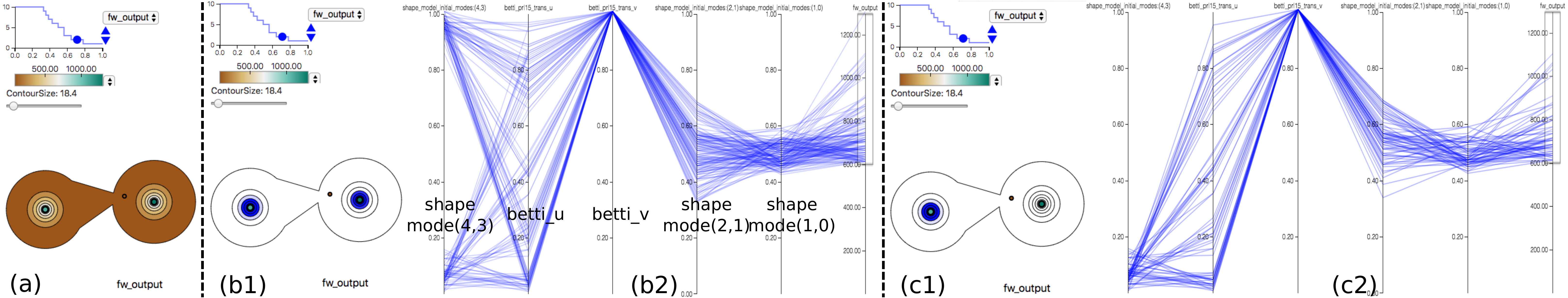}
  \vspace{-3mm}
 \caption{Joint exploration of both topological and geometric characteristics of the surrogate's errors as functions in the input parameter space. In (a1), the topological spine shows two local peaks of the output error ($\Ferror_o(x)$). The user can select the high error samples through parallel coordinates plot in (b2). These samples will also be highlighted in topological spine (b1). Also, the user can focus on an individual extremum (see c1, c2).
}
\label{fig:ICF_fw}
\vspace{-3mm}
\end{figure*}

\subsection{Deep Learning for Inertial Confinement Fusion}
Although describing all the architectural details of the model is beyond the scope of the paper, Fig.~\ref{fig:cycleGAN} illustrates the key components.
In this study, the input to the simulation code is a 5D
parameter space $\Xspace \subset \Rspace$.  For each combination of
input parameters $x_i\in\Xspace$, the simulation code produces 12 mulitchannel images
$\mathbf{I}_j =\{I_j^0, ..., I_j^{11}\}$, each of size
$64 \times 64 \times 4$ in the image space $\Ispace$, as well as 15 scalar quantities
$\mathbf{y}_j = \{y_j^0, ..., y_j^{14}\} \in \Sspace = \Rspace^{15}$.
To jointly predict the images and scalars, the model uses a
bow-tie autoencoder~\cite{Tolstikhin2017} ($\Ispace \times \Sspace \rightarrow \Lspace \rightarrow \Ispace \times \Sspace$, not shown in the figure) to construct a joint latent space
$\Lspace \subset \Rspace^{20}$ that captures all image and scalar variations.
The forward model $\Xspace \rightarrow \Ispace \times \Sspace$ is comprised of
two components: a multivariate regressor
$F: \Xspace \rightarrow \Lspace$ and the decoder from the pretrained autoencoder: $D: \Lspace \rightarrow \Ispace \times \Sspace$.
In other words, the forward model does not directly predict the system output $\Ispace \times \Sspace$,
but instead predicts the joint latent space $\Lspace$ of the autoencoder.
The decoder $D: \Lspace \rightarrow \Ispace \times \Sspace$ can then produce the actual outputs from the predicted $\Lspace$.
Such a setup allows us to more effectively utilize relationships in the output domain $\Ispace \times \Sspace$
and improve the overall predictive performance.
Furthermore, the model simultaneously considers the inverse
model $G: \Lspace \rightarrow \Xspace$ and uses self-consistency to
regularize the mapping, by minimizing $||x - G(F(x))||_2$ in addition
to the prediction loss. All submodels are implemented using deep
neural networks (DNNs), and the entire system is trained jointly to
obtain the fitted models $\hat F$ and $\hat G$, assuming that we have access to a pretrained autoencoder.
In order to enable neural network training at such large scales, and to support the system integration needs, we adopted the parallel neural network training toolkit, LBANN~\cite{van2015lbann}, for training the surrogate.

\subsection{Surrogate Model Analysis}
Although the ability of the model to produce accurate and self-consistent
predictions of multimodal outputs provides fundamentally new
capabilities for scientific applications, the complexity of the resulting model naturally leads to challenges in its evaluation and exploration.
Firstly, the application requirement for scientific deep learning is vastly different from that of its commercial counterparts.
Compared to traditional applications such as image recognition and text processing (where a human user can easily provide the ground truth), in scientific problems, even an expert has limited knowledge about the behavior of an experiment, due to the inherent complexity of the physical system as well as the exploratory nature of the domain.
Secondly, these models are intended for precise and quantitative experimental designs that, we hope, can lead to scientific discovery.
Consequently, physicists care about not only the overall prediction accuracy,
but also the localized behaviors of a model, i.e., whether certain regions of the input parameter space produce more error than others.

To address these challenges, we include the proposed visualization tool as an integral part of the model design/training process.
As discussed in Section~\ref{sec:task}, the ability to interpret a high-dimensional function is central for obtaining the localized understanding of a given system. In this application, the functions of interest are the high-dimensional error landscapes of the surrogate model. By adopting the proposed function visualization tool, we aim to provide the model designers and the physicists with intuitive feedback on the prediction error distribution in the input parameter space. 
For over five months, we have worked closely with the machine learning team to iteratively debug and fine-tune the training process.
We have detected a crucial issue with the normalization strategy applied to the x-ray images and identified the problem with the batch scheduler based on the visualization provided by the proposed tool. 
The following analysis is carried out on a ``well-behaved'' model, where the aforementioned problems have already been addressed, according to the traditional evaluation metrics (e.g., average prediction accuracy, loss convergence behavior).
Here, we illustrate how the exploration of localized error behaviors can reveal some important yet unexpected issues of the surrogate model that are not possible with the traditional summary metrics.

\para{Exploring surrogate error in the input parameter space}
Traditionally, the machine learning community has relied on global
summary statistics to assess a model, i.e., global loss curves, that do not reveal localized properties.
By utilizing the proposed tool, we view the predicative error as a function in the input domain,
which can then be analyzed as a high-dimensional scalar function (see \textbf{T2-4} in Section~\ref{sec:task}).
Such a line of inquiry is not only essential for scientific analysis but also can provide a critical feedback loop for validating and fine-tuning the actual machine learning model.
In order to perform an elaborate as well as unbiased study, we carry out our analysis on an 8 million validation set, hold-out during the training process.
Despite the large sample size, due to the scalable design considerations, we are able to show that the proposed system allows an interactive linked-view exploration once the topological structure and the corresponding datacube are obtained.
The precomputation of the 8 million sample dataset took around 1.5-3.5 hours to complete, depending on the initial neighborhood query size as well as the hardware setup (i.e., utilizing GPU or not). 


During the model training process, a global average loss function is considered, which in this case consists
of a weighted sum of the losses in $\Lspace$ (forward model), $\Sspace$
(decoder), and $\Xspace$ (inverse model). 
Each term addresses a different aspect of the overall training objectives.
More specifically, we are given:
(1) the autoencoder reconstruction error, $\Rerror(l) = || y - D(l)||$, for
$l \in \Lspace$ and $l = E(y)$ denoting the encoder of the autoencoder;
(2) the forward error in latent space $\Ferror_{lat}(x) = ||\hat F(x) - E(y)||$, where $\hat F(x)$ is the fitted forward model;
(3) the forward error in output space $\Ferror_o(x) = || y - D(\hat F(x)) ||$;
and (4) the cyclic self-consistency error $\Gerror(x) = ||x - \hat G(\hat F(x))||$.

To gauge the overall error of the surrogate, let us first look at the forward error in output space $\Ferror_o(x)$ (i.e., the error in the predicted images and diagnostic scalars of the surrogate).
We compute the topological structure of the error in the input parameter space.
As shown in Fig.~\ref{fig:ICF_fw} (a), from the topological spine, we can see that there are two distinct local extrema of high errors in the 5D parameter space.
We highlight the relationships between those samples through an aggregated parallel coordinate plot shown in (b2). We can also equivalently show where these samples are in the topological spine as shown in (b1). Note, a different shade of blue is used to indicate the fraction of the selected samples associated with each contour of the ``topological terrain''. A darker shade indicates a larger fraction.
To understand the relationship between the patterns in the PCP and the peaks in the topological spine, we can focus on one of the peaks in the topology spine (c1), which will also trigger an update of the PCP. As shown in (c2), the left peak appears to correspond to samples with low values in the first dimension (shape\_mode\_(4, 3)), indicating that the two peaks are maximally separated in that direction. 


\begin{figure*}[!t]
\vspace{-2mm}
\centering
  \includegraphics[width=0.97\linewidth]{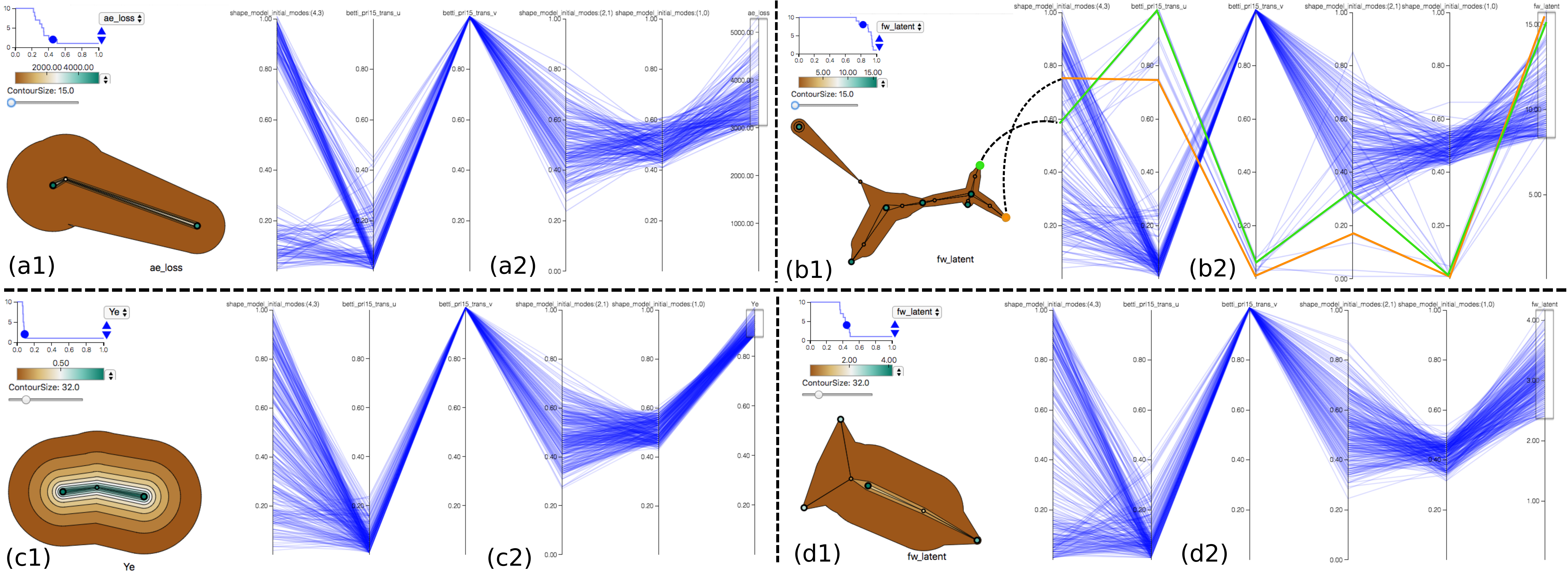}
  \vspace{-3mm}
 \caption{
The autoencoder error ($\Rerror(l)$) and the latent space error ($\Ferror_{lat}(x)$) is shown in (a),(b) respectively. The yield of the simulation is shown in (c). In (d), we illustrate the latent space error ($\Ferror_{lat}(x)$) of the model that are trained for 80 epochs instead of 40.
}
\label{fig:ICF_comparison}
\vspace{-4mm}
\end{figure*}

\para{Interaction between different types of errors}
The error in the output space $\Ferror_o(x)$ is affected by both forward error in the latent space $\Ferror_{lat}(x)$ and the autoencoder error in $\Rerror(l)$.
Hence, we subsequently examine all three error components side by side.
Let us first look at the error in the autoencoder, which is trained separately from the forward/inverse models.
As shown in Fig.~\ref{fig:ICF_comparison}(a1), interestingly, the spine also contains two peaks, and they are maximally separated by the first dimension (Fig.~\ref{fig:ICF_comparison}(a2)), which is similar to the topology of forward error in the output space (Fig.~\ref{fig:ICF_fw}(b1, b2)). 
However, the error in latent space $\Ferror_{lat}(x)$, as illustrated in Fig.~\ref{fig:ICF_comparison}(b1), has a much more complex, yet stable, topological structure.
When  we focus on an individual topological segment (in Fig.~\ref{fig:ICF_comparison}(b1, b2), we use orange and green to highlight two of the extrema (b1) and their corresponding lines in the parallel coordinate (b2)). We notice the extrema corresponds to outlying patterns in the parallel coordinate plot that will be often ignored by typical statistical analysis techniques. By viewing all three error patterns together, we find that (1) despite similar overall similarity between the PCP plots, the majority of the local extrema of $\Ferror_{lat}(x)$ (such as the one highlighted in (b1,b2)) do not reappear in the $\Ferror_o(x)$ (see Fig.~\ref{fig:ICF_fw}); (2) the $\Rerror(l)$ and $\Ferror_o(x)$ have a similar diverging pattern in the first dimension (see PCP plots), which is not found in $\Ferror_{lat}(x)$.
Such an observation seems to indicate that the autoencoder error has a very strong influence on the output error of the surrogate.
As a result, it is important to further explore the potential cause of the high error in autoencoder.
In addition, the forward error in the latent space $\Ferror_{lat}(x)$ has some very out-of-ordinary local extrema (b2) - it would also be interesting to understand the cause (is the error due to faulty outputs in the simulation, or are the image features more challenging to predict, e.g., high-frequency content) and potentially fix the errors.

\noindent\textbf{What contributes to the high error?}
Upon examination of the behavior of the prediction error in the input domain, we are curious to understand the possible causes for the observed patterns (\textbf{T3}).
Interestingly, from the PCP plot of the simulated energy yield (see Fig.~\ref{fig:ICF_comparison}(c2)), we find patterns (especially in the last three dimensions) similar to that of the previously explored error components.
This interesting discovery could have a significant impact on the application since the physicists are interested in the transition regions between low and higher yield.
Despite the similarity between the high yield and high error in the autoencoder (see Fig.~\ref{fig:ICF_comparison}(a2, c2), there is a clear discrepancy in the first dimension (i.e., \emph{shape\_model\_(4, 3)}), where the autoencoder error exhibits a clear binary pattern (higher error corresponds to high or low values). 
Apparently, then, another factor besides the yield is correlated with the autoencoder error.

\begin{figure}[htbp]
\vspace{-2mm}
\centering
  \includegraphics[width=1.0\linewidth]{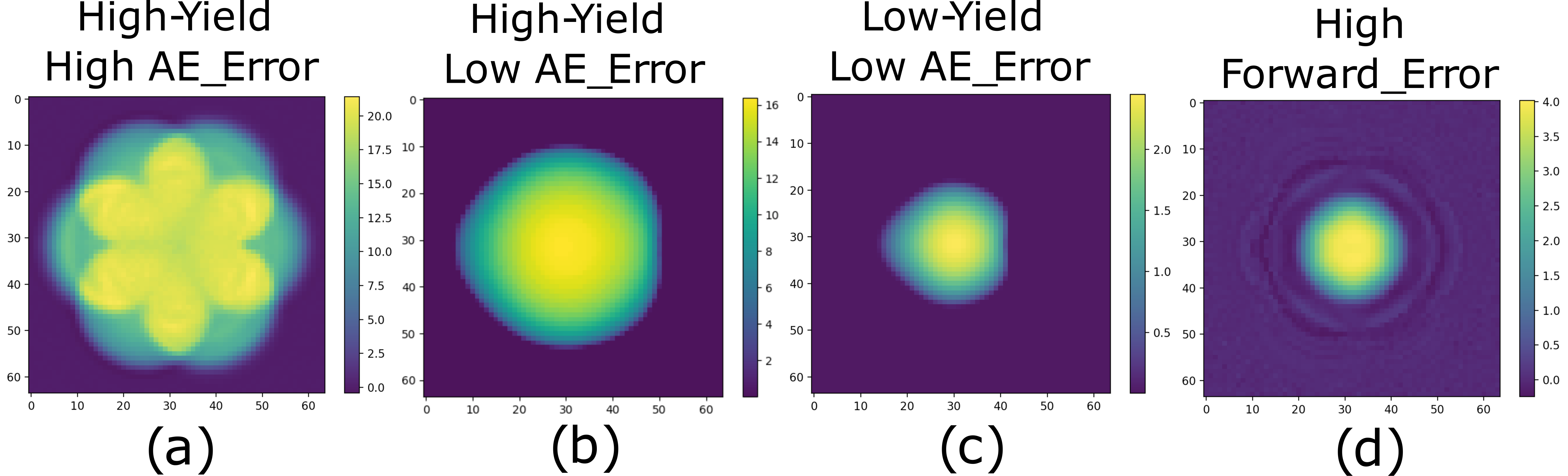}
  \vspace{-6mm}
 \caption{
	X-ray images with different energy and error conditions.
 }
\label{fig:ICF_image}
\vspace{-2mm}
\end{figure}

We need to examine the difference between x-ray images corresponding to high/low yield and high/low error conditions, as the autoencoder error directly indicates how well the model can reconstruct the images and scalars from the latent representations.
As shown in Fig.~\ref{fig:ICF_image}, we identify two images (a, b) with high and low autoencoder error, respectively.
Although both correspond to high yield conditions, image (a) has a much more complex pattern, which is more challenging to reconstruct, thus leading to  a larger autoencoder error when compared to the image (b).
According to the physicists, the first parameter (\emph{shape\_model\_(4, 3)}) encodes the higher order shape information. As a result, the simulator is expected to generate image patterns similar to (b) when the parameter value is closer to $0$. Larger deviation from $0$ induces more complex patterns, as we observe in the image (a).

\noindent\textbf{How many samples do we need to train the surrogate?}
As noted previously in Fig.~\ref{fig:ICF_comparison} (b2), a number of local extrema in different parts of the domain are characterized by a high forward model error $\Ferror_{lat}(x)$. To analyze this behavior, we identify the corresponding images (see Fig.~\ref{fig:ICF_image}(d)), which exhibit high-frequency patterns around the central circular regions, making it harder to predict. The natural question is, does this mean we need more training data to better learn these patterns or did we not train the model until convergence?

To answer these questions, we carried out experiments by varying the training size and training time. By increasing the training time from 40 to 80 epochs, we observe that the out-of-ordinary local extrema in Fig.~\ref{fig:ICF_comparison}(b2) disappear (see Fig.~\ref{fig:ICF_comparison}(d2)), which indicates that the model had not converged sufficiently.
This result was quite surprising since the average loss curve appeared to have reached a steady state, long before the 40 epochs, and we still chose to continue training. 
Based on feedback from the physicists, we realized that the ``cloud'' around the high-density center may not reflect the true physics, and could be an artifact of the simulator or the image rendering process. 
This discovery is crucial because it is not otherwise feasible to examine these large datasets for such anomalous behaviors.

\begin{figure}[htbp]
\centering
 \vspace{-2mm}
  \includegraphics[width=1.0\linewidth]{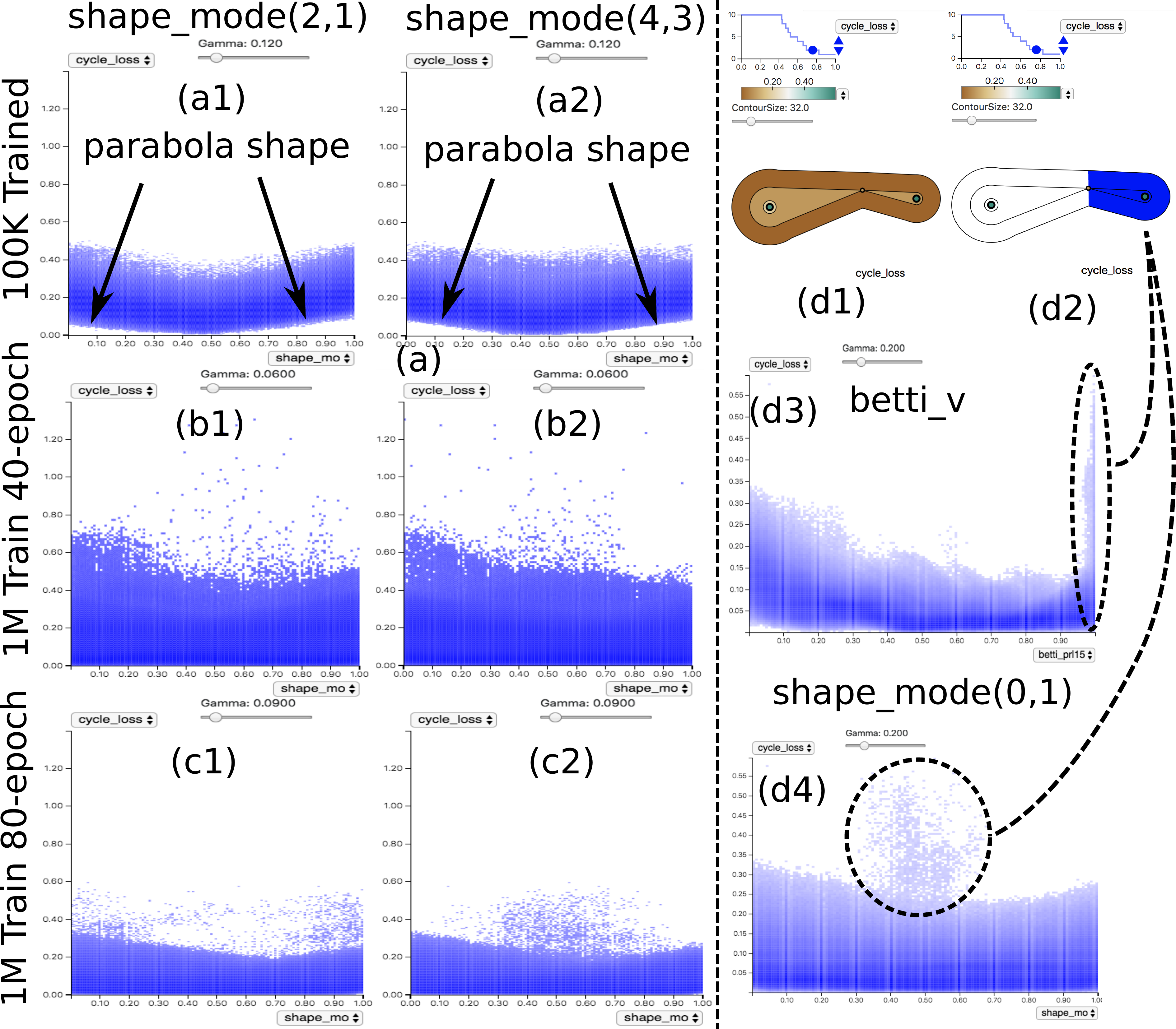}
  \vspace{-6mm}
 \caption{The comparison of cyclic loss (a measure of surrogate self-consistency) among models with different training sets and training times. 
As shown in (a), the 100K samples trained model exhibits a peculiar parabola pattern, which indicates the lack of self-consistency for a certain parameter range.
By increasing the training set (from 100K to 1 Million), we are able to eliminate these empty regions (see (b)), but the model ends up with larger errors, especially among outliers, which can be mitigated by increasing the training time (from 40 to 80 epochs, see (c)).
In (d), we show the topological structure of the self-consistency loss of the best model and reveal that the highly inconsistent region is characterized by high values of physical parameter \emph{betti\_v} of the surrogate input.
} 
\label{fig:ICF_cyclic_comparison}
\vspace{-2mm}
\end{figure}

Similar to the training time, we found that changes to training size also drastically affect error behaviors. In Fig.~\ref{fig:ICF_cyclic_comparison}, we show a comparison of the cyclical errors in three different training setups (in the plot, the y-axis is the error, and the x-axis is marked for each column at the top of the figure).
The cyclic error $\Gerror(x)$ provides a way to enforce/evaluate model self-consistency, which is crucial for building physically meaningful models.
For the model trained using 100k samples (Fig.~\ref{fig:ICF_cyclic_comparison}(a)), we can see the error exhibits a rather peculiar parabolic shape along certain dimensions (see (a1, a2)). The empty region in the error plots reveals that for certain parameter combinations, the model will always be inconsistent to a certain extent, which is not desirable. However, by increasing the training set to 1 million samples (even without increasing the training epochs), as shown in Fig.~\ref{fig:ICF_cyclic_comparison}(b), we no longer see those empty regions in the parameter space. However, with more data, the overall errors do not improve; in fact, the outliers become more severe (all plots on the left panel have the same y-axis scale). As expected, we could reduce the discrepancies by increasing the training duration (Fig.~\ref{fig:ICF_cyclic_comparison}(c)).
The deep learning experts postulate that certain modes of the images (i.e., more complex ones) likely require larger number examples to learn, which the 100k training samples cannot provide.
Finally, in Fig.~\ref{fig:ICF_cyclic_comparison}(d), we show the topological structure of the cyclic error of the (1M training samples, 80-epoch) model,
where one of the peaks corresponds to a high-error region characterized by large values for one of the physics parameters \emph{betti\_v}. As noted by our collaborators, the instability of the surrogate is most likely due to the volatile nature of physics around the point of implosion, where the diagnostic quantities can change drastically.

The above analysis for the ICF surrogate model clearly demonstrates that scientific applications require a new suite of evaluation strategies, based on both statistical characterization and exploratory analysis. The insights provided by our approach cannot be obtained otherwise by using aggregated statistics of errors.
For analyzing similar surrogate models, as a general guideline, we believe it is crucial to first explore the relationship between the input parameter space and the corresponding localized errors. We can then dive into the contributors of the overall error and examine their causes to infer unintended behaviors of the model. 
Lastly, to fully evaluate the model, we also need to understand how the error distribution in the input space impacts the target application. 
For example, a high concentration of localized error may not always indicate an undesirable model, as the application may require precision only where the model is highly accurate. 

\section{Application: Multiscale Simulation for Cancer Research}
Although surrogate modeling is a natural use case for our approach, high-dimensional functions appear in many other applications, sometimes in unexpected contexts.
Here, we discuss another ongoing collaboration with computational biologists, who are interested in 
understanding a particular type of cancer-signaling chain through both experimental and computational means.
Specifically, they focus on analyzing how the RAS protein interacts with the human cell membrane to start a cascade of 
signals, leading to cell growth. 
Most types of aggressive-cancers, such as pancreatic, are known to be linked to RAS mutations that cause unregulated, 
i.e., cancerous, growth of the affected cells. 
However, so far, the attempts to affect RAS function through drugs have led to complete disruption of signals and 
have ultimately proven fatal to the patient.


The cell membrane is made up of two layers of lipids 
that constantly move, driven by complex dynamics dependent on lipid type, surrounding proteins, local curvatures, etc. 
Scientists suspect that the local lipid composition underneath the RAS protein plays a major role in the signaling cascade. 
Unfortunately, the relevant length scales are not accessible experimentally.
Therefore, molecular dynamics (MD) simulations are the only source of information. 
Using the latest generation of computer models and resources, it is now possible to study the interaction of the 
RAS protein with the cell membrane for both the atomistic and the so-called coarse-grained MD simulations.
However, such simulations are costly and typically restricted to micro-second intervals at nano-meter scales. 
The challenge is that at these time and length scales, seeing changes in RAS configuration or unusual lipid mixtures is extremely rare, and thus the chances of simulating even one event of interest using a pseudo-random setup are very low. 
Instead, scientists are interested in simulating micro-meter-sized lipid bilayers for up to seconds of time, which would not only make events of interest more likely, but also reach experimentally observable scales, 
which is crucial to calibrate and verify computational results. 
To address these seemingly contradictory requirements, our collaborators are developing a multiresolution 
simulation framework (Fig.~\ref{fig:workflow}). 
At the coarse level, a continuum simulation is used that abstracts individual lipids into spatial concentrations and models RAS proteins as individual particles. 
Such simulations are capable of reaching biologically relevant length and timescales and are expected to provide insights 
into local lipid concentrations as well as clustering of RAS proteins on the cell membrane.
However, such models cannot deliver an analysis of the molecular interactions.
For such detailed insights, MD simulations are performed using much smaller subsets of the bilayer in regions that are considered ``interesting.''
The key challenge is to determine which set of fine-scale simulations must be executed to provide the greatest 
chances for new insights.

\subsection{Analysis of Importance Sampling}
\begin{figure}[htbp]
\centering
\includegraphics[width=\linewidth]{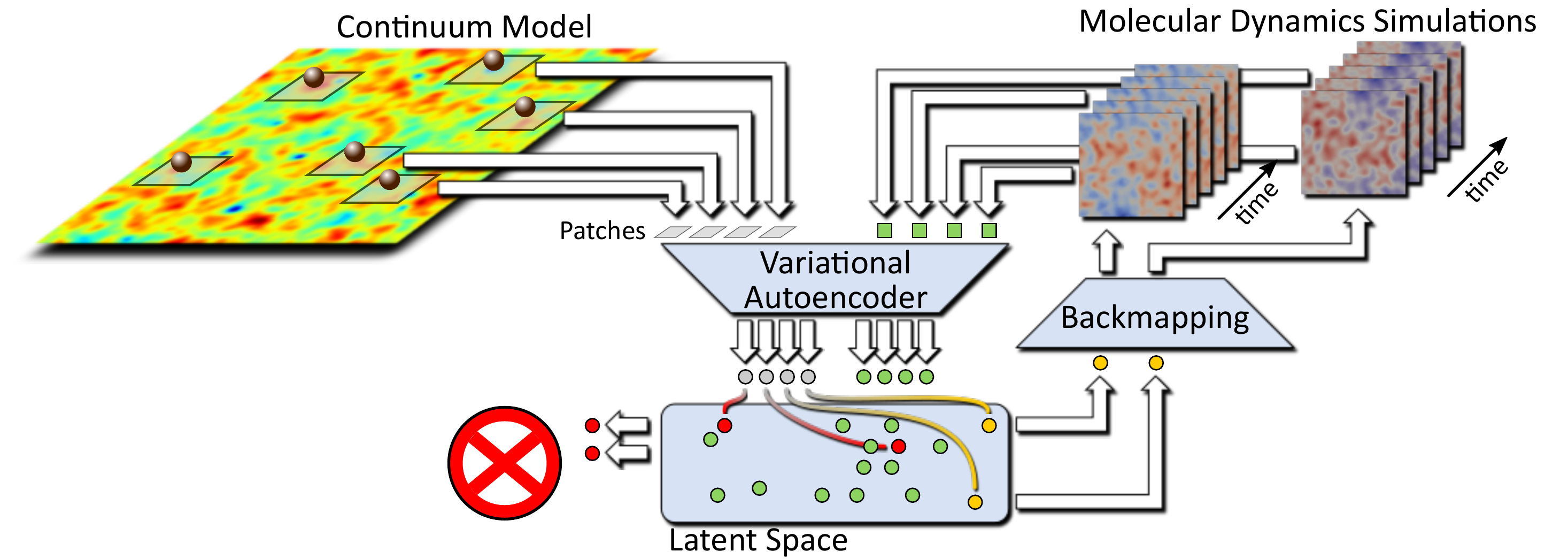}
\vspace{-4mm}
\caption{Multiscale simulation of RAS-membrane biology for cancer research.
As the continuum model (left) evolves, {\it patches} around the RAS proteins (shaded balls) are projected into a high-dimensional latent space, which is used to drive an importance sampling. Patches that are different from previous simulations (yellow) are used to initialize new MD simulations, whereas patches too close to existing samples (red) are discarded. 
Significant challenges exist in the analysis of high-dimensional latent spaces to improve the sampling.}
\label{fig:workflow}
\vspace{-2mm}
\end{figure}

As shown in Fig.~\ref{fig:workflow}, the current approach uses an autoencoder-style setup 
similar to the one discussed in the previous section. 
Each local \emph{patch} underneath a RAS protein in the continuum simulation is projected into a 15D latent space formed by the bottleneck layer of an autoencoder. 
In the context of this paper, the latent space is a nonlinear dimensionality reduction of the space of all patches onto 15 dimensions. 
Given limited computational resources, the goal is to determine which of these patches must be examined more 
closely, using MD simulations. 
To this end, our collaborators are using a farthest-point sampling approach in the latent space. As resources become available, 
they choose the patches (yellow) that are farthest away in the latent space from all previously selected patches (green).
The intuition is that this strategy, in the limit, will evenly sample the space of all patches rather than repeatedly executing common configurations. 
In this manner, it will provide maximal information for the available computing resources and drive the process toward rare configurations.
In this context, ``rare'' refers to the simulation length and timescales, which are still very small, i.e., seconds and micrometers scale.
The practical challenge is to determine how well this sampling strategy is working and what its practical effects are 
(see \textbf{T1} in Section~\ref{sec:task}). 
Initial attempts used nonlinear embeddings, e.g., t-SNE, to compare a hypothetical random sample to the optimized farthest-point sampling.
Due to the inherent nonlinear distortion, however, the results were inconclusive and difficult to interpret. 
Other straightforward approaches, e.g., comparing distribution of neighborhood distances between random and optimized samples, also provided little insight. 

\begin{figure*}[!t]
\vspace{-1mm}
\centering
  \includegraphics[width=.99\linewidth]{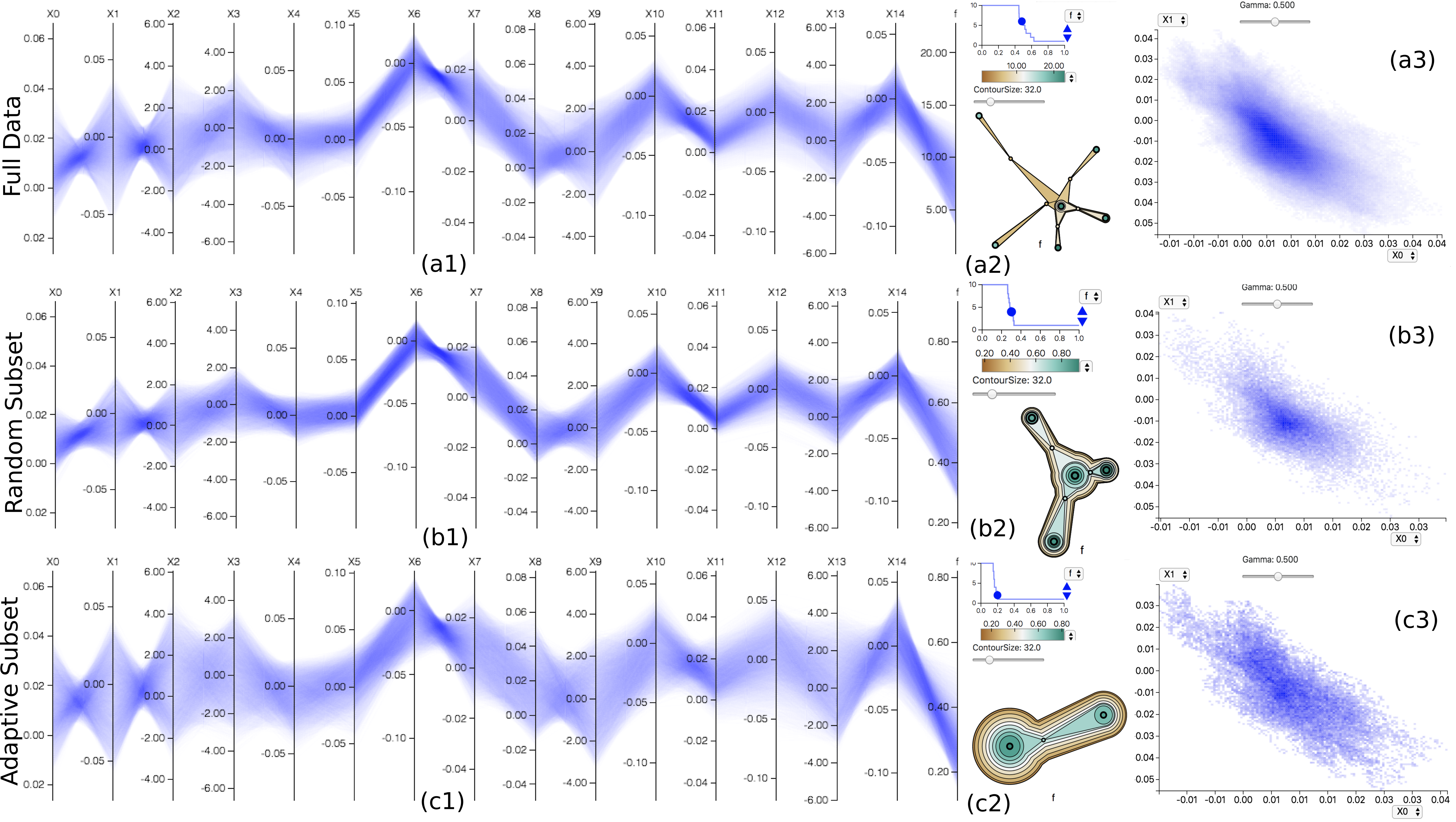}
\vspace{-3mm}
\caption{Visualization of different sampling patterns (middle and bottom) when 
compared to the overall distribution (top) highlights that the adaptive importance-based approach 
has a wider coverage and lower and fewer modes, as required to execute the targeted multiscale simulation.}
\label{fig:pilot2}
\vspace{-3mm}
\end{figure*}


We use our tool to explore these sampling patterns.
In particular, our collaborators estimate the point density in the latent space as the inverse of the mean 
distance to 20 nearest neighbors. The density is computed for three sets of points: 
1) all available patches (2 million points); 2) a uniform random subselection of all patches (100K points); 
and 3) an optimized subselection using farthest-point sampling (100K points). Given this density 
estimate, we compute the extrema graph for all three sets.
As shown on the top row of Fig.~\ref{fig:pilot2}, the set of all patches results in multiple high-persistence plateaus (a2), which indicates well-separated density maxima, the smallest of which has a persistence of $\sim 50\%$.
Note that these modes are not apparent in the parallel coordinates (a1) nor in the scatterplot of the latent dimension ((a3) shows the first two latent dimensions). 
The mid-row plots show the random selection, which creates fewer modes (b2) with lower persistence $\sim 30\%$, 
which are also significantly less stable.
The figure shows four major modes for illustration, but a more pragmatic interpretation is that there exists only a single mode of density, and the remaining maxima are due to noise, 
which is surprising since the random sample is expected to mimic the full distribution. 
In fact, our collaborators typically rely on analyzing the random subset assuming their equivalence. 
One current hypothesis is that the modes are simply too small to be reliably sampled with 100K points. 
Another possibility is that the high-dimensional density estimation introduces artifacts. 
The bottom-row plots show the result of the adaptive importance-based sampling with even smaller modes and lower overall density peaks.
Furthermore, the parallel coordinates show a substantially wider distribution, even though the highest density peak ($\sim 0.82$) remains, 
which is surprisingly similar to the one of the random sample ($\sim 0.95$).
Again, the similarity is likely a consequence of the density estimation (which applies a nonlinear scaling factor). 
Nevertheless, our analysis and visual representations have provided direct and intuitive evidence to our collaborators 
regarding the effectiveness of their sampling strategy, which was not apparent in their previous analyses (e.g., tSNE plots). 
The team has executed this sampling as part of a large multiscale simulation using 
a parallel workflow for several days utilizing all 176,000 CPUs and 16,000 GPUs of 
\emph{Sierra}, the second-fastest supercomputer in the world, aggregating over 116,000 MD simulations.


\section{Discussion and Future Work}
In this work, we have identified a set of common tasks often encountered in analyzing data derived from computational pipelines for scientific discovery and introduced a scalable visual analytic tool to address these challenges via a joint exploration of both topological and geometric features.
To achieve the analysis objectives and facilitate large-scale applications, we employed a streaming construction of extremum graphs and introduced the concept of topological-aware datacube to aggregate large datasets according to their topological structure for interactive query and analysis.
We highlight the fact that scientific deep learning requires a different set of evaluation and diagnostic schemes due to the drastically different objectives.
In scientific applications, it is often not sufficient to obtain the best average performance or identify the typical simulation results. 
Instead, it is more important to provide insight and establish confidence in the model itself and understand where or why the model may be unreliable for real-world scenarios.
As demonstrated in the application, the proposed tool helps us not only evaluate and finetune the surrogate model but also identify a potential issue in the physical simulation code that would otherwise be omitted.
We believe the application-aware error landscape analysis demonstrated in this work is both valuable and necessary for many similar deep learning applications in the scientific domain.

For future work, we plan to further exploit the parallelism in the neighborhood query process by partitioning the problem domain and then merging the query results on the fly. Also, we plan to extend the current topology-aware datacube to support online queries of a wide range of joint distribution representations (e.g., parametric density model, copulas) and incorporate efficient compression strategies.
Finally, we are in the process of releasing the proposed tool as part of a general purpose high-dimensional data analysis package to better facilitate the analysis and evaluation of similar applications.

\acknowledgments{
This work was performed under the auspices of the U.S. Department of Energy by Lawrence Livermore National Laboratory under Contract DE-AC52-07NA27344. The work was also supported in part by NSF:CGV Award: 1314896, NSF:IIP Award: 1602127, NSF:ACI Award:1649923, DOE/SciDAC DESC0007446, PSAAP CCMSC DE-NA0002375 and NSF:OAC Award: 1842042. Additional support comes from Intel Graphics and Visualization Institutes of XeLLENCE program. 

}

\bibliographystyle{abbrv-doi-hyperref}

\bibliography{topoBibtex}

\begin{thebibliography}{10}

\bibitem{jagData}
Jag icf dataset for scientific machine learning.
\newblock \url{https://github.com/rushilanirudh/icf-jag-cycleGAN}.
\newblock Accessed: 2019-07-15.

\bibitem{anderson1958introduction}
T.~W. Anderson.
\newblock {\em An introduction to multivariate statistical analysis}, vol.~2.
\newblock Wiley New York, 1958.

\bibitem{artero2004uncovering}
A.~O. Artero, M.~C.~F. de~Oliveira, and H.~Levkowitz.
\newblock Uncovering clusters in crowded parallel coordinates visualizations.
\newblock In {\em |}, pp. 81--88. IEEE, 2004.

\bibitem{baldi2001bioinformatics}
P.~Baldi, S.~Brunak, and F.~Bach.
\newblock {\em Bioinformatics: the machine learning approach}.
\newblock MIT press, 2001.

\bibitem{bennett2012combining}
J.~C. Bennett, H.~Abbasi, P.-T. Bremer, R.~Grout, A.~Gyulassy, T.~Jin,
  S.~Klasky, H.~Kolla, M.~Parashar, V.~Pascucci, et~al.
\newblock Combining in-situ and in-transit processing to enable extreme-scale
  scientific analysis.
\newblock In {\em Proceedings of the International Conference on High
  Performance Computing, Networking, Storage and Analysis}, p.~49. IEEE
  Computer Society Press, 2012.

\bibitem{nddav}
P.-T. Bremer, D.~Maljovec, A.~Saha, B.~Wang, J.~Gaffney, B.~K. Spears, and
  V.~Pascucci.
\newblock Nddav: N-dimensional data analysis and visualization analysis for the
  national ignition campaign.
\newblock {\em Computing and Visualization in Science}, 17(1):1--18, 2015.

\bibitem{Bremer06Banff}
P.-T. Bremer, V.~Pascucci, and B.~Hamann.
\newblock Maximizing adaptivity in hierarchical topological models using
  cancellation trees.
\newblock In T.~Moeller, B.~Hamann, and B.~Russell, eds., {\em Mathematical
  Foundations of Scientific Visualization, Computer Graphics, and Massive Data
  Exploration}, p. to appear. Springer, 2006.

\bibitem{butler2018machine}
K.~T. Butler, D.~W. Davies, H.~Cartwright, O.~Isayev, and A.~Walsh.
\newblock Machine learning for molecular and materials science.
\newblock {\em Nature}, 559(7715):547, 2018.

\bibitem{CorreaLindstrom2011}
C.~Correa and P.~Lindstrom.
\newblock Towards robust topology of sparsely sampled data.
\newblock {\em IEEE Transactions on Visualization and Computer Graphics},
  17(12):1852--1861, Dec. 2011. doi: {{%
10\hspace{.1pt}\discretionary{.}{%
}{.}\hspace{.4pt}1109\discretionary{/}{%
}{/}TVCG\hspace{.1pt}\discretionary{.}{%
}{.}\hspace{.4pt}2011\hspace{.1pt}\discretionary{.}{%
}{.}\hspace{.4pt}245}}


\bibitem{CorreaLindstromBremer2011}
C.~Correa, P.~Lindstrom, and P.-T. Bremer.
\newblock Topological spines: A structure-preserving visual representation of
  scalar fields.
\newblock {\em IEEE Transactions on Visualization and Computer Graphics},
  17(12):1842--1851, Dec. 2011. doi: {{%
10\hspace{.1pt}\discretionary{.}{%
}{.}\hspace{.4pt}1109\discretionary{/}{%
}{/}TVCG\hspace{.1pt}\discretionary{.}{%
}{.}\hspace{.4pt}2011\hspace{.1pt}\discretionary{.}{%
}{.}\hspace{.4pt}244}}


\bibitem{dang2010stacking}
T.~N. Dang, L.~Wilkinson, and A.~Anand.
\newblock Stacking graphic elements to avoid over-plotting.
\newblock {\em IEEE Transactions on Visualization and Computer Graphics},
  16(6):1044--1052, 2010.

\bibitem{gaffney2014thermodynamic}
J.~Gaffney, P.~Springer, and G.~Collins.
\newblock Thermodynamic modeling of uncertainties in nif icf implosions due to
  underlying microphysics models.
\newblock In {\em APS Meeting Abstracts}, 2014.

\bibitem{gerber2010visual}
S.~Gerber, P.-T. Bremer, V.~Pascucci, and R.~Whitaker.
\newblock Visual exploration of high dimensional scalar functions.
\newblock {\em IEEE transactions on visualization and computer graphics},
  16(6):1271, 2010.

\bibitem{Gerber10tvcg}
S.~Gerber, P.-T. Bremer, V.~Pascucci, and R.~Whitaker.
\newblock Visual exploration of high dimensional scalar functions.
\newblock {\em {IEEE} Transactions on Visualization and Computer Graphics},
  16(6):1271--1280, 2010.

\bibitem{gyulassy2008practical}
A.~Gyulassy, P.-T. Bremer, B.~Hamann, and V.~Pascucci.
\newblock A practical approach to morse-smale complex computation: Scalability
  and generality.
\newblock {\em IEEE Transactions on Visualization and Computer Graphics},
  14(6), 2008.

\bibitem{gyulassy2005topology}
A.~Gyulassy and V.~Natarajan.
\newblock Topology-based simplification for feature extraction from 3d scalar
  fields.
\newblock In {\em Visualization, 2005. VIS 05. IEEE}, pp. 535--542. IEEE, 2005.

\bibitem{KrausePererNg2016}
J.~Krause, A.~Perer, and K.~Ng.
\newblock Interacting with predictions: Visual inspection of black-box machine
  learning models.
\newblock In {\em Proceedings of the 2016 CHI Conference on Human Factors in
  Computing Systems}, pp. 5686--5697. ACM, 2016.

\bibitem{landge2014situ}
A.~G. Landge, V.~Pascucci, A.~Gyulassy, J.~C. Bennett, H.~Kolla, J.~Chen, and
  P.-T. Bremer.
\newblock In-situ feature extraction of large scale combustion simulations
  using segmented merge trees.
\newblock In {\em High Performance Computing, Networking, Storage and Analysis,
  SC14: International Conference for}, pp. 1020--1031. IEEE, 2014.

\bibitem{lins2013nanocubes}
L.~Lins, J.~T. Klosowski, and C.~Scheidegger.
\newblock Nanocubes for real-time exploration of spatiotemporal datasets.
\newblock {\em IEEE Transactions on Visualization and Computer Graphics},
  19(12):2456--2465, 2013.

\bibitem{liu2018analyzing}
M.~Liu, S.~Liu, H.~Su, K.~Cao, and J.~Zhu.
\newblock Analyzing the noise robustness of deep neural networks.
\newblock {\em arXiv preprint arXiv:1810.03913}, 2018.

\bibitem{Liu18usc}
S.~Liu, K.~Humbird, L.~Peterson, J.~Thiagarajan, B.~Spears, and P.-T. Bremer.
\newblock Topology-driven analysis and exploration of high-dimensional models.
\newblock In {\em Research Challenges and Opportunities at the interface of
  Machine Learning and Uncertainty Quantification}, 2018.

\bibitem{liu2019nlize}
S.~Liu, Z.~Li, T.~Li, V.~Srikumar, V.~Pascucci, and P.-T. Bremer.
\newblock Nlize: A perturbation-driven visual interrogation tool for analyzing
  and interpreting natural language inference models.
\newblock {\em IEEE transactions on visualization and computer graphics},
  25(1):651--660, 2019.

\bibitem{liu2013immens}
Z.~Liu, B.~Jiang, and J.~Heer.
\newblock immens: Real-time visual querying of big data.
\newblock In {\em Computer Graphics Forum}, vol.~32, pp. 421--430. Wiley Online
  Library, 2013.

\bibitem{LundbergLee2017}
S.~M. Lundberg and S.-I. Lee.
\newblock A unified approach to interpreting model predictions.
\newblock In {\em Advances in Neural Information Processing Systems}, pp.
  4768--4777, 2017.

\bibitem{mayorga2013splatterplots}
A.~Mayorga and M.~Gleicher.
\newblock Splatterplots: Overcoming overdraw in scatter plots.
\newblock {\em IEEE transactions on visualization and computer graphics},
  19(9):1526--1538, 2013.

\bibitem{ming2019rulematrix}
Y.~Ming, H.~Qu, and E.~Bertini.
\newblock Rulematrix: Visualizing and understanding classifiers with rules.
\newblock {\em IEEE transactions on visualization and computer graphics},
  25(1):342--352, 2019.

\bibitem{mjolsness2001machine}
E.~Mjolsness and D.~DeCoste.
\newblock Machine learning for science: state of the art and future prospects.
\newblock {\em science}, 293(5537):2051--2055, 2001.

\bibitem{OlahMordvintsevSchubert2017}
C.~Olah, A.~Mordvintsev, and L.~Schubert.
\newblock Feature visualization.
\newblock {\em Distill}, 2017.
\newblock https://distill.pub/2017/feature-visualization. doi: {{%
10\hspace{.1pt}\discretionary{.}{%
}{.}\hspace{.4pt}23915\discretionary{/}{%
}{/}distill\hspace{.1pt}\discretionary{.}{%
}{.}\hspace{.4pt}00007}}


\bibitem{peterson2017zonal}
J.~Peterson, K.~Humbird, J.~Field, S.~Brandon, S.~Langer, R.~Nora, B.~Spears,
  and P.~Springer.
\newblock Zonal flow generation in inertial confinement fusion implosions.
\newblock {\em Physics of Plasmas}, 24(3):032702, 2017.

\bibitem{RibeiroSinghGuestrin2016}
M.~T. Ribeiro, S.~Singh, and C.~Guestrin.
\newblock Why should i trust you?: Explaining the predictions of any
  classifier.
\newblock In {\em Proceedings of the 22nd ACM SIGKDD International Conference
  on Knowledge Discovery and Data Mining}, pp. 1135--1144. ACM, 2016.

\bibitem{SimonyanVedaldiZisserman2013}
K.~Simonyan, A.~Vedaldi, and A.~Zisserman.
\newblock Deep inside convolutional networks: Visualising image classification
  models and saliency maps.
\newblock {\em arXiv preprint arXiv:1312.6034}, 2013.

\bibitem{springer2013integrated}
P.~Springer, C.~Cerjan, R.~Betti, J.~Caggiano, M.~Edwards, J.~Frenje, V.~Y.
  Glebov, S.~Glenzer, S.~Glenn, N.~Izumi, et~al.
\newblock Integrated thermodynamic model for ignition target performance.
\newblock In {\em EPJ Web of Conferences}, vol.~59, p. 04001. EDP Sciences,
  2013.

\bibitem{Thomas13tvcg}
D.~M. Thomas and V.~Natarajan.
\newblock Detecting symmetry in scalar fields using augmented extremum graphs.
\newblock {\em IEEE Transactions on Visualization and Computer Graphics},
  19(12):2663--2672, Dec 2013. doi: {{%
10\hspace{.1pt}\discretionary{.}{%
}{.}\hspace{.4pt}1109\discretionary{/}{%
}{/}TVCG\hspace{.1pt}\discretionary{.}{%
}{.}\hspace{.4pt}2013\hspace{.1pt}\discretionary{.}{%
}{.}\hspace{.4pt}148}}


\bibitem{Tolstikhin2017}
I.~Tolstikhin, O.~Bousquet, S.~Gelly, and B.~Schoelkopf.
\newblock Wasserstein auto-encoders.
\newblock {\em arXiv preprint arXiv:1711.01558}, 2017.

\bibitem{van2015lbann}
B.~Van~Essen, H.~Kim, R.~Pearce, K.~Boakye, and B.~Chen.
\newblock Lbann: Livermore big artificial neural network hpc toolkit.
\newblock In {\em Proceedings of the Workshop on Machine Learning in
  High-Performance Computing Environments}, p.~5. ACM, 2015.

\bibitem{wang2019dqnviz}
J.~Wang, L.~Gou, H.-W. Shen, and H.~Yang.
\newblock Dqnviz: A visual analytics approach to understand deep q-networks.
\newblock {\em IEEE transactions on visualization and computer graphics},
  25(1):288--298, 2019.

\bibitem{YosinskiCluneNguyen2015}
J.~Yosinski, J.~Clune, A.~Nguyen, T.~Fuchs, and H.~Lipson.
\newblock Understanding neural networks through deep visualization.
\newblock {\em arXiv preprint arXiv:1506.06579}, 2015.

\bibitem{ZeilerFergus2014}
M.~D. Zeiler and R.~Fergus.
\newblock Visualizing and understanding convolutional networks.
\newblock In {\em European conference on computer vision}, pp. 818--833.
  Springer, 2014.

\end{thebibliography}
\end{document}